# Efficient Distributed Training through Gradient Compression with Sparsification and Quantization Techniques

**Shruti Singh[1], Shantanu Kumar[2]**

sshruti.connect@gmail.com[1], reach.shantanuk@gmail.com[2]

[1] Washington State University, Seattle, WA, USA

[2] Amazon, Seattle, WA, USA

*Abstract*— **This study investigates the impact of gradient compression on distributed training performance, focusing on sparsification and quantization techniques, including top-k, DGC, and QSGD. In baseline experiments, random-k compression results in severe performance degradation, highlighting its inefficacy. In contrast, using top-k and DGC at 50× compression yields performance improvements, reducing perplexity by up to 0.06 compared to baseline. Experiments across 1, 2, and 4 workers demonstrate that conservative sparsification can have a regularizing effect, especially for smaller models, while compression ratios above 5000× impair performance, particularly for DGC. Communication times are reduced across all compression methods, with top-k and DGC decreasing communication to negligible levels at high compression ratios. However, increased computation times offset this efficiency for top-k due to sorting demands, making it less scalable than DGC or QSGD. In convergence tests, sparsification techniques show accelerated convergence, requiring fewer epochs than the baseline, which has implications for computational savings. Although precision trade-offs among sparsification, floating point errors are mitigated by compression. This study's findings underscore the need to tune hyperparameters specifically for each compression technique to achieve optimal model performance, especially in distributed training systems.**

*Index Terms*— **DL, Distributed, Gradient Compression, Resources, Security**

## I. INTRODUCTION

Deep Learning (DL) has established itself as a powerful methodology for tackling a diverse array of tasks, including but not limited to machine translation [1], [2], [3], speech recognition [4], [5], and object detection [6], [7]. The rapid advancement of this field is attributed to several key factors: the increasing availability and volume of multi-modal data for training complex models [8], [9], significant innovations in Machine Learning (ML) algorithms and techniques—such as Convolutional Neural Networks (CNNs) [10] and the seminal "ImageNet moment"—and, most critically, the expanded access to more cost-effective and powerful hardware. The integration of Graphical Processing Units (GPUs) in DL has played a pivotal role in facilitating the training of deeper and more sophisticated models [11]. GPUs excel in performing highly parallel computations, thereby substantially reducing training time. This scaling of model size and computational power is evident in the continuous improvements seen across various domains, with notable examples including the advancements in parameter-rich language models. However, this scaling also introduces increased training costs. For instance, training widely used deep models such as ResNet-50 on a single GPU can take several days. To mitigate this challenge of scale, distributed DL systems have been introduced as a mechanism to parallelize model training [12], [13], [14], [15]. These systems utilize large-scale clusters of machines, or workers, to distribute data batches evenly, allowing computations to be executed in parallel and subsequently aggregated into a global model update. While this approach reduces the training time for complex models, it also brings about additional costs related to communication. Networks of machines must frequently communicate individual updates, a process that is often constrained by the bandwidth limitations of the interconnects. Over the past decade, while GPUs have experienced performance speedups of up to five times, network bandwidth in distributed systems has not seen proportional improvements [16], [17]. This growing disparity between the computational power of GPUs and communication bandwidth has created a situation where model parameter updates are generated more rapidly than they can be transmitted. This bottleneck can overwhelm central parameter servers in distributed DL architectures, leading to underutilization of costly GPU resources as they remain idle during communication delays. As a result, addressing this communication bottleneck has become a pressing research area in distributed DL systems.

Therefore, in the domain of gradient compression [18], a wide array of factors can be adjusted and explored, including model architecture, model size, dataset, optimizer, and hyperparameters. This broad scope of potential modifications complicates the comparison of novel gradient compression techniques due to the lack of consistent testing environments and frameworks. A notable issue arises when comparing two prominent techniques in gradient quantization [19]—a specific form of compression—where conclusions are drawn despite differences in the model architectures employed. Furthermore, some techniques are introduced without accounting for critical metrics, such as computational overhead, which can lead to a misrepresentation of the technique's true effectiveness. There is a scarcity of comprehensive surveys that directly compare gradient compression techniques, and those that do exist tend to be predominantly qualitative. For practitioners seeking to implement gradient compression to address the challenges of distributed systems, the current literature often fails to provide a practical reference to help determine the most appropriate technique for their specific needs [20], [21]. Consequently, the burden falls on the practitioner to conduct time-consuming and resource-intensive experiments. This study identifies an additional gap in the literature related to gradient compression. Most surveys, whether qualitative or quantitative, focus primarily on the field of computer vision [22], [23], [24], [25]. In contrast, Natural Language Processing (NLP) models typically exhibit a lower computation-to-communication ratio than computer vision models, which often benefit from weight-sharing patterns and parallel computation [26]. Given that gradient compression aims to mitigate communication bottlenecks in distributed systems, focusing on Recurrent Neural Network (RNN) architectures—common in NLP tasks—presents a valuable opportunity for evaluation. These architectures, with their lower computation-to-communication ratio, have been less thoroughly investigated in existing literature. In consideration of these points, the objectives of this study are as follows:

1. RO1: Conduct a comparative analysis of selected gradient compression techniques from the literature, utilizing a standardized framework that ensures fair and accurate comparisons, enabling robust conclusions.
2. RO2: Evaluate all relevant metrics associated with distributed model training using gradient compression, including test accuracy, convergence rates, computational time, and compression ratio.
3. RO3: Focus on RNN architectures, providing an evaluation of gradient compression techniques in a context that is both highly relevant and underexplored in the current body of research.

Therefore, in this study, it is hypothesized that each gradient compression technique will effectively compress gradients while maintaining test performance that does not substantially diverge from the baseline case of zero compression. Consistent with prevailing research trends, it is anticipated that the Deep Gradient Compression (DGC) [18] technique will exhibit particularly superior performance as compression ratios are scaled to larger values. However, it is also foreseen that this enhanced performance may be accompanied by trade-offs in terms of increased operational costs, especially as distributed environments grow in both the number of workers and the size of the models. This complexity introduces ambiguity in selecting the optimal compression technique under such circumstances.

The study is as follows; the theoretical backdrop will be seen in the following section. The related works are presented in Section III. The examination of the compression model is done in Section IV. The experimental analysis is carried out in Section V. The result analysis is given in Section VI, and we wrap up the paper with some conclusions and ideas for future work in Section VII.

## II. THEORETICAL BACKGROUND

### A. Distributed Deep Learning

Distributed DL refers to an infrastructure composed of multiple compute nodes that can process data in parallel [12], [13], [14], [15], collaborating to train Deep Neural Networks (DNNs) models. These compute nodes typically incorporate multiple GPUs, or workers, utilized to expedite the training and inference of DL models. Additionally, other hardware components, such as Field-Programmable Gate Arrays (FPGAs) and Google Tensor Processing Units (TPUs), may be employed depending on the specific application requirements. Recent surveys of network architectures indicate a prevailing preference for GPUs [27], particularly emphasizing the increasing significance of multi-node architectures in research, which reflects the growing importance of distributed learning for accelerating experimentation. To facilitate communication among compute nodes, the most commonly employed interconnection is the InfiniBand[1] adapter, specifically designed for use in distributed DL environments involving GPU node clusters. While Ethernet[2] is another popular interconnect that offers high bandwidth, Infiniband provides lower latency, making it more suitable for rapid communication. Nevertheless, advancements in network communication technology have not kept pace with the progress in hardware accelerators. Consequently, the high costs associated with 100 Gbps+ connection infrastructures are frequently overlooked in favor of more economical yet slower alternatives, such as widely used Amazon EC2 data center instances that operate at up to 25 Gbps. In addition to these hardware infrastructure considerations, several other critical decisions must be made when designing distributed systems. Below, three such decision areas are outlined, each playing a significant role in shaping communication-efficient distributed DL and informing the design choices discussed in Section V.

#### 1) Parallelism

Parallelism refers to the methodology by which tasks are divided among computational resources [28]. The most prevalent approach is data parallelism [29], wherein the dataset is partitioned into non-overlapping batches that are fed into identical models residing on each worker within a distributed system. In this paradigm, the majority of operations—excluding certain processes such as batch normalization—during the feed-forward and backpropagation steps of a deep network are executed on an individual sample level. This characteristic enables the independent execution of mini-batch parameter updates, which are subsequently aggregated at the conclusion of each full iteration. However, an alternative technique is model parallelism [30], which distributes computational workload across the neurons within each layer, effectively segmenting the model itself. This method is often employed in situations where a model exceeds the capacity of a single GPU's memory. However, model parallelism presents two significant challenges: first, determining the optimal allocation of splits is a complex problem that has been shown to be NP-complete[3]; second, partitioning by structural components results in increased penalties associated with the volume of information that must be communicated between compute nodes. As a result, data parallelism is more commonly adopted in practical applications.

---

[1] https://www.nvidia.com/en-in/networking/infiniband-adapters/
[2] Devices can connect via data packets on Local Area Networks (LANs) thanks to Ethernet, a wired networking technology.

### 2) Centralisation

Centralization in distributed environments refers to the architectural setup governing the communication and coordination among compute nodes [31]. This setup generally falls into two primary categories. The first is a centralized regime, in which updates from multiple workers are transmitted to a single central location, typically a parameter server. These updates are then applied simultaneously to a master model, with the resulting parameter changes communicated back to each worker. The second category is a decentralized regime, which relies on protocols such as Message Passing Interface (MPI) to facilitate communication directly between the nodes themselves, often through methods like *allreduce* or *allgather*. The choice between centralized and decentralized architectures depends on various factors, including network bandwidth, topology, and latency. Decentralized architectures are often viewed as a preferable option since they do not require additional resources such as parameter servers and eliminate the risk of a single point of failure within the network. More relevant to this work is the aspect of communication efficiency. Decentralized architectures offer logarithmic scaling in communication with respect to the number of workers, which reduces overhead. Additionally, they circumvent the need for "double compression", a process in centralized regimes where gradient compression must be performed in both directions, further adding to computational and communication costs.

### 3) Synchronization

Synchronization refers to the degree of coordination required between communicated updates in distributed learning systems [32]. In synchronous updates, global aggregation and model modifications must be completed and communicated across all nodes before the next iteration can proceed. This approach mirrors the structure of classic Stochastic Gradient Descent (SGD), which allows for convergence properties to be more rigorously analyzed. However, a key drawback of synchronous updates is that the overall process is limited by the performance of the slowest worker in the network, often referred to as the "straggler" effect. In contrast, asynchronous updates allow messages to be sent as soon as they are ready, alleviating the delay caused by slower workers. However, this flexibility introduces the issue of "stale" parameters, where gradient updates may become less relevant as they are based on outdated information from previous iterations. This can negatively impact the convergence properties of the model. Due to these concerns, synchronous updates are generally the more widely adopted method in distributed training environments.

### B. Communication Efficient Training

Addressing the challenge of communication in distributed DL is crucial. With computational power advancing at a significantly faster rate than network connection speeds in distributed systems, the number of updates required per unit time often exceeds current bandwidth capabilities. Two prominent approaches have been developed to mitigate this issue: i) Communication Scheduling [33]: In widely-used DL frameworks, the timing of parameter update communication is often random, leading to high variance in both iteration and communication times. This randomness can result in inefficiencies, making it essential to optimize the timing and sequence of updates in order to maximize overall system performance. ii) Gradient Compression [18]: Another key method for enhancing communication efficiency is through the compression of model parameters, specifically gradients. One widely-adopted technique in this category involves reducing the precision of model parameters from the standard 32-bit floating point representation commonly used in DL. This reduction in precision can yield significant savings in communication volume, often more than doubling efficiency. This work, however, focuses on gradient compression, which minimizes the volume of gradient updates communicated between workers during the optimization process. Gradient compression techniques can be broadly classified into two categories: gradient quantization [19] and gradient sparsification [34]. These methods aim to reduce the communication load while maintaining the performance and accuracy of the model, thereby addressing the communication bottleneck in distributed DL environments.

### 1) Gradient Quantization

Gradient quantization is a compression technique designed to reduce the bit representation of the components of a gradient vector by mapping continuous values to discrete levels or "buckets" [19]. The earliest form of quantization reduced values to 16-bit representations, achieving a compression ratio of 2× since each component occupies half the memory footprint. Numerous quantization algorithms have been developed since, and it is worth noting that due to inherent limitations of the method, the maximum achievable compression ratio is approximately 32×. Limited-bit quantization techniques, which truncate the number of bits used to represent the gradient, follow a trend of increasing compression ratios. These techniques range from 8-bit quantization to the most extreme case, 1-bit quantization. The latter can take various forms, such as a threshold-based scheme, where values below a certain threshold are compressed to '0' and others to '1', or a sign-based scheme that encodes values based solely on their sign. Due to the lossy nature of these techniques, it is crucial to implement an error-feedback mechanism to ensure model convergence. This mechanism retains the quantization error from the compression in memory and adds it back to the corresponding gradient during the next iteration, helping maintain convergence properties. QSGD introduces a probabilistic [21], stochastic rounding function that defines a family of algorithms, enabling users to balance the trade-off between compression ratio and variance in convergence guarantees by tuning the size of the buckets used. Further quantization techniques include adaptive quantization [35], which dynamically adjusts to the gradients to provide a non-static compression ratio, and split-gradient quantization [36], which compresses gradient differences to help ensure model convergence properties. These innovations reflect ongoing efforts to optimize communication efficiency in distributed DL without compromising the accuracy or convergence of the model.

---

[3] A class of computer problems that are both in NP and as difficult as any other NP problem is known as NP-complete.

### 2) Gradient Sparsification

The other prominent form of gradient compression is sparsification [34], a technique that selectively transmits a portion of the gradient components, sending only their values and corresponding indices (at full 32-bit precision). The first algorithm to introduce sparsification was truncated gradient [37], was initially aimed at addressing memory and computational constraints during model training. However, sparsification has an added advantage over quantization in terms of reducing the communication volume, as it is not confined to the $32\times$ compression ratio limitation observed in quantization methods. Further investigations into gradient updates in DNNs reveal a distribution of values that are narrowly dispersed and often close to zero, making sparsification particularly well-suited for compression. Fundamental approaches to gradient sparsification typically involve masking or thresholding, which determines the components that will be transmitted during communication between workers. One of the simplest methods is random-$k$ [38], where $k$ % of gradient values are randomly selected for transmission. Although this method is somewhat rudimentary, more sophisticated techniques like top-$k$ are widely used [38]. In Top-$k$, gradient values are ordered by their absolute magnitude, and only the top $k$ % are transmitted, ensuring that the most impactful gradients—those with the largest magnitudes—are sent. Other methods propose using fixed thresholds, above which gradient components are transmitted, though these thresholds can be challenging to fine-tune. As with quantization, an error-feedback mechanism is often incorporated in sparsification to carry over any residual errors to subsequent iterations, improving convergence. The Top-$k$ selection method is further enhanced by DGC [18], which incorporates momentum and other techniques to improve the error-feedback process and mitigate the staleness often observed in such correction steps. Other sparsification techniques employ adaptive thresholding approaches, such as AdaComp [39], which adjusts the level of sparsification according to the local activity of each layer, making compression more effective across different layer types. Sketched-SGD [38] estimates the Top-$k$ gradients using a Count-Min sketch data structure, which offers additional benefits in terms of the size and scalability of the communicated updates. The literature on gradient compression is rich with methods that go beyond standalone quantization and sparsification. A comprehensive review of these techniques can be found in a dedicated survey [40].

### C. Language Modelling

The present study centers on gradient compression within the context of language modeling. This focus aligns with the third research objective: to explore models that are inherently more communication-intensive. By doing so, this work aims to complement prior research efforts by providing a comprehensive and detailed comparison of gradient compression techniques specifically tailored for language modeling tasks. Language modelling defines a task that sets out to estimate the joint probability distribution over a sequence of tokens [41]. In the early stages of language modeling research, the context was constrained to a fixed size under the Markov[4] assumption. This approach, however, encountered significant challenges, particularly regarding long-range token dependencies and computational complexity. More recently, continuous conditional language models, such as RNNs [42], have addressed these issues by employing co-learned word representations. In this framework, tokens are embedded and subsequently transmitted to a hidden layer that models the complete historical context before being projected back onto a comprehensive vocabulary matrix for softmax activation. This study specifically focuses on the Long Short-Term Memory (LSTM) [43] variant of RNNs, which incorporates a 'memory cell' mechanism designed to enhance the encoding of long-term dependencies in token context. Language models are predominantly assessed using a metric known as perplexity, which fundamentally represents the weighted average branching factor of a language model. Through this definition, we understand it as the typical number of feasible words a language model will 'choose' between-the lower the better.

### III. RELATED WORKS

Quantizing gradients to low-precision values can significantly reduce communication bandwidth. Previous studies have explored various approaches to achieve this, including 1-bit SGD [20], QSGD [21], and TernGrad [44]. These methods have demonstrated the convergence of quantized training, with TernGrad focusing on CNNs and QSGD examining the training loss of RNNs. Further research has investigated quantizing the entire model, including gradients. For instance, DoReFa-Net [45] employs 1-bit weights with 2-bit gradients, while [46] proposed threshold quantization. However, the choice of threshold poses practical challenges. To address this, [47] selected a fixed proportion of positive and negative gradient updates, and [48] introduced gradient dropping, which sparsifies gradients using a single threshold based on absolute value. Gradient dropping achieves 99% gradient exchange savings with only 0.3% loss of BLEU score on machine translation tasks. Concurrently, [39] developed a method to automatically adjust the compression rate based on local gradient activity, achieving compression ratios of up to $200\times$ for fully-connected layers and $40\times$ for convolutional layers with negligible degradation of top-1 accuracy on the ImageNet dataset. In contrast, DGC achieves a compression ratio of up to $600\times$ for the entire model without requiring extra layer normalization or modifying the model structure. Most importantly, DGC results in no loss of accuracy.

---

[4] Future system states only depend on the current state and not on previous ones, according to the Markov assumption.

## IV. Pre-Implementations: Compression Model Analysis

This section provides a detailed overview of the gradient compression techniques explored in this study. For each technique, a comprehensive background is presented, along with the rationale for its selection over alternative methods, in alignment with research objective 1. Random-$k$ is a gradient sparsification technique that selects a fixed proportion $k$, of the components from the gradient update vector, retaining only this subset of gradient elements for communication during the SGD updates. In contrast, top-$k$ employs a more informed strategy, ordering the gradient vector elements by their absolute magnitude and selecting the top proportion based on this ranking. In both approaches, any elements not selected for communication are set to zero, thereby reducing the overall volume of data being transmitted. To enhance the convergence and generalization properties of these techniques, error feedback, referred to as residual memory in this work, is employed [38]. This method stores the difference between the compressed and uncompressed gradients prior to communication, tracking the compression errors and allowing them to be added back to the gradients in subsequent iterations. This mechanism ensures that smaller gradient elements eventually contribute to parameter updates, as they grow in magnitude through successive residual memory updates. When calculating the compression ratio for sparsification techniques, it is important to note that both the values of the gradient elements and their corresponding indices must be communicated to allow for decompression at the receiving end. This effectively reduces the perceived compression ratio by half. For instance, a Top-0.01 method, which retains 1% of the gradient elements, achieves a compression ratio of $100/2 = 50\times$. Although more sophisticated methods for packing sparse updates exist, such as DGC, this work adopts the value-index communication method, as it remains the most commonly used approach. Both gradient compression techniques—random-$k$ and top-$k$—are fundamental and widely referenced in the literature. Random-$k$ serves as a simple, naive baseline, useful for benchmarking more advanced methods. Top-$k$, on the other hand, offers a higher compression ratio and is efficient to implement, making it a practical choice for practitioners seeking straightforward yet effective solutions. Each technique leverages built-in library functions to perform tensor manipulations. Random-$k$ utilizes the *torch.randperm()* function, while top-$k$ employs the *torch.topk()* function. It is worth noting that sorting the tensor directly may offer faster performance on GPUs in practice. Consequently, this approach is also explored during the experimentation phase to assess its potential benefits.

DGC [18] employs gradient sparsification to achieve high compression ratios, ranging from $270\times$ to $600\times$, with minimal impact on test perplexity. DGC enhances the traditional top-$k$ error-feedback mechanism by improving the local accumulation of gradients and addressing the issue of staleness in the residual errors fed back to gradients. It accomplishes this through the application of three key techniques: i) Momentum Correction: It modifies the momentum-based SGD formula, enabling momentum to be applied to residual memory gradient updates, as used in methods like top-$k$. This adjustment ensures that even small gradient updates, which would otherwise be neglected, are appropriately accumulated and corrected over time, contributing to more effective model convergence and preserving accuracy despite the high compression rates. This correction process refines the trajectory of parallel SGD, contributing to improved model convergence when sparse updates are employed. The momentum parameter is considered a tunable hyperparameter in this study, allowing for optimization to enhance the effectiveness of the gradient compression approach. ii) Momentum Factor Masking: It is employed to address momentum staleness in residual gradients. This staleness occurs because the correction updates for specific parameters may require thousands of iterations to accumulate sufficiently before being applied during communication. Consequently, gradients risk being directed in a stale trajectory that no longer reflects the current state of the model. By applying this mask, the momentum of affected gradient components is set to zero, preventing them from continuing in an outdated direction. iii) Local Gradient Clipping: It is performed at each node rather than after gradient aggregation, given that gradients accumulated on individual nodes are independent. Warm-up training, which involves using a smaller learning rate during the early stages of training, is also utilized to mitigate the impact of delayed gradient updates, which may initially be aggressive. In terms of communication, the DGC employ an encoding scheme that preserves knowledge of the gradient tensor's shape, transmitting values as 32-bits and the lengths of zero-runs between values as 16-bits, rather than transmitting both values and indices. While this approach is more efficient, such differences in implementation may complicate fair comparisons between compression techniques. In this work, a simpler approach is assumed, with naive communication of both values and indices.

Whereas, Quantised-SGD (QSGD) [21] represents a family of gradient compression techniques that aim to reduce the bit representation of gradient values. This reduction is accomplished through the use of randomized rounding, which stochastically assigns gradient values to a set of discrete quantisation levels. By adjusting the number of available quantisation levels, QSGD allows for a trade-off between the communication time required for distributed gradient updates and the variance introduced into the model. Consequently, this trade-off also affects the model's convergence guarantees, providing flexibility in balancing compression efficiency and model performance. The use of stochastic rounding ensures that the resulting quantised gradient provides an unbiased estimate of the original gradient component. In addition to stochastic rounding, the QSGD employ Elias[5] encoding to pack values losslessly into a lower bit representation. This encoding method is particularly efficient since smaller magnitude numbers require fewer bits to encode, and larger values, which occur less frequently in data, consume more bits. However, Elias coding is not applied in this study, as is the case with other quantitative surveys [49]. Similar to the exclusion of efficient packing observed with DGC, this omission ensures that all compression techniques are evaluated on a comparable basis. However, due to the relatively limited compression ratios that quantisation techniques offer, and within the constraints of this study, only one quantisation method has been selected. QSGD was chosen because of its ability to provide a variable compression ratio, enabling a detailed analysis of the trade-offs between model performance and compression ratio in the context of quantisations.

---

[5] Elias encoding is a universal coding system that uses variable-length prefix codes to efficiently encode positive numbers.

## V. EXPERIMENTAL ANALYSIS

### A. Experimental Setup

In terms of computation, running experiments with models of significant size (i.e., those with over 1 million parameters) necessitates access to a GPU. It is important that the computational resource used is a dedicated machine, ensuring stability in the computation time metric across different runs. This requirement ruled out the use of shared GPU cluster resources or platforms such as Google Colab, which often allocate shared GPU memory between multiple users. For this study, a virtual machine instance was established on the Google Cloud Platform's free trial services. The selected setup utilizes an NVIDIA T4 GPU with 16GB of memory, providing approximately 30 days of computation time through the use of free credits. Each model, when run using the fair random search method has an average runtime of approximately four hours, which constrains the number of experiments that can be executed within the allocated time. Consequently, the decision was made to evaluate a smaller number of compression methods but to investigate a broader range of variables, including model size, the number of workers, and the compression ratios employed by the selected compression techniques.

To ensure a fair comparison of results, a novel procedure for hyperparameter tuning is implemented. Unlike the common practice of retaining hyperparameters tuned for the baseline model, this approach performs a random search over the hyperparameter space for each model type, including variations in model size and compression techniques. This procedure is repeated for 60 iterations, which has a 95% chance of finding a combination of parameters within 5% of the optimal values. Early stopping with model checkpointing is also implemented on the validation set to save training time and analyze convergence properties. Additionally, plans are made to run repeat experiments on the best-case hyperparameters for each model type to estimate error bounds. However, due to significant performance variations observed even with different random seeds, this procedure is limited to experiments run on a single seed (42). In line with background research, this work assumes synchronous, data-parallel training and estimates communication times based on a parameter server architecture with bi-directional communication. This formal approach ensures a rigorous and fair comparison of results across different model types and techniques. Each individual experiment in this study focuses on varying three key experimental variables to examine the effects of modifying and scaling common distributed settings on model performance. These variables are adjusted within a predetermined set of options, allowing the creation of a comprehensive landscape of results: i) Number of Workers: This variable represents the (simulated) number of GPUs used for training the distributed model. The baseline scenario involves a single worker, and the scaling of this variable is explored within the set $\{1, 2, 4\}$. This enables an assessment of how performance changes as the number of workers increases. ii) Model Size: This variable corresponds to the number of units in the hidden layers of the LSTM model, which is varied between 100 (small), 300 (medium), and 500 (large). Adjusting this parameter changes the number of trainable parameters, with values ranging from 1,161,600 to 9,008,000 parameters. This variation allows for an analysis of scalability in relation to different model sizes. Notably, these sizes are significantly smaller than the large hidden layer sizes commonly found in the literature (~1500) [18]. iii) Compression Ratio: The compression ratio is modified for each compression technique to evaluate the limits of its effectiveness. For sparsification techniques, such as top-$k$ and DGC, the ratio is varied on a logarithmic scale within the set $\{0.01, 0.001, 0.0001\}$. These values correspond to 50×, 500×, and 5000× compression, respectively, when considering the communication of both gradient values and indices. For QSGD, the compression ratio is determined by the quantum number variable, which is tested with values $\{64, 16, 4\}$, resulting in 7-bit, 5-bit, and 3-bit compression, respectively. In terms of compression ratio, this translates to 4.6×, 6.4×, and 10.7× compression, respectively.

Random searches are conducted using the weights and biases experiment tracking software, specifically employing the sweeps[6] feature. During hyperparameter sweeps, the learning rate is drawn from a uniform distribution, U~[1,50], with the optimal range of values observed to be higher than expected, likely due to the use of vanilla SGD. The dropout rate is sampled from U~[0,0.8]. For experiments utilizing DGC, the momentum hyperparameter is selected from U~[0.1,0.9]. The batch size is fixed at 256, selected to maximize GPU memory usage, as monitored through the *nvidia-smi* command, and evenly distributed among workers in a distributed setting. In all experiments, weights are tied between the input and output embeddings. When applying early stopping, a patience parameter of 10 is used, and models are trained for a maximum of 50 epochs, as fewer epochs were required for convergence during baseline testing. All experiments assume the use of vanilla SGD, with alternative optimizers reserved for future investigations. Additionally, no learning rate decay is applied, as its absence was found to yield better results. For the LSTM model parameters, a sequence length (or BPTT parameter) of 35 is consistently used, along with a fixed architecture of two *torch.nn.LSTM* layers. This decision was made to minimize the complexity in the number of variables explored during experimentation.

---

[6] With the sweeps function, we may systematically investigate the impact of changing parameters in simulations or optimizations across a predetermined range.

## B. Datasets

In alignment with research objective 3, this study focuses on the language modeling sub-task, employing the word-level language modeling Penn Treebank (PTB) dataset [50]. The PTB dataset comprises 923,000 training tokens, 73,000 validation tokens, and 82,000 test tokens, all of which have been pre-processed. Due to its manageable size and extensive use in comparative language modeling surveys, it is a popular choice for this type of research. Moreover, its relatively smaller size compared to other widely-used datasets makes it ideal for the experimental resources allocated for this study. During data processing, a vocabulary size of 10,000 is assumed, with all less frequent tokens represented by the [UNK] token. The focus on language models, rather than CNNs—typically a more common testbed for compression techniques—is justified by the lower computation-to-compression ratio of language models in distributed systems [18]. The model employed in this research is a LSTM network. Despite the growing prominence of transformer-based language models, the LSTM family was selected for two principal reasons: first, the majority of gradient-based language model compression research concentrates on LSTM models, thus aligning this study with existing literature. Second, transformer-based models generally require significantly more computational resources due to their larger parameter scale, which exceeds the available computational capacity of this study. Expanding the analysis of compression techniques to include other types of language models is a potential avenue for future research. The LSTM model used in this study adheres to a standard architecture, while incorporating several features identified in the literature, such as those from the widely recognized AWD-LSTM[7] architecture. The AWD-LSTM, however, is not fully implemented here for two primary reasons: first, by focusing on a more basic model, the analysis and conclusions regarding gradient compression techniques can be drawn without the influence of additional regularization methods specific to AWD-LSTM. Second, the novel simulated environment introduced in this study is easier to implement and debug with a simpler model structure. One feature from AWD-LSTM that is incorporated is weight tying between the input and output embedding layers, which reduces the number of trainable parameters. This reduction in parameters allows for the training of larger, more complex models within the study's computational constraints. Additionally, the input and output dimensions of the first and last layers of the LSTM are fixed to the embedding dimension size, further reducing the overall parameter count.

## C. Distributed Environment Setup

Given the limited resources available for this study, a novel approach is adopted to simulate the effects of gradient compression in distributed systems on a single machine (GPU). In data-parallel distributed training, each worker computes on its own partition of the mini-batch at each iteration, making updates to the global loss independent of other workers. As such, it is feasible to simulate the distributed learning process through the technique of gradient accumulation [51]. Gradient accumulation allows for multiple backward passes through the neural network before parameter gradients are zeroed out, as is conventionally done. Similar to data-parallel training, the full batch of data is processed in mini-batches. However, in this simulation, the mini-batches are processed sequentially on the same machine. The gradients from each mini-batch are accumulated, cumulatively forming the final full-batch gradient. Once all gradients from the mini-batches are calculated and accumulated, an optimizer update step is applied. This method effectively simulates distributed training, where separate workers (with identical models) process data splits in parallel. In this study, a custom optimizer was developed to accommodate gradient compression within this framework. In a standard gradient accumulation process, the mini-batch gradients for each parameter are added directly into the *grad()* variable. However, in this simulation, each gradient must be compressed before accumulation. This is achieved by copying the parameter gradient, compressing it, and then adding it to a local variable that will be used for the optimizer update step. This approach maintains the ability to zero out the parameter gradients between iterations, as is typically done. The custom optimizer extends the capabilities of the Horovod[8] distributed DL framework. Simulating distributed model training on a single GPU offers the advantage of assessing distributed model performance within a resource-constrained environment, as is the case in this study. However, the serial nature of mini-batch processing, as opposed to parallel processing, presents a drawback in that computation time is higher compared to a fully distributed setting. Furthermore, the absence of practical steps such as *allreduce* communication reduces the accuracy of performance metrics such as communication time. These limitations are considered when interpreting the results.

## D. Evaluation Metrics

To ensure a fair evaluation of each compression method, as aligned with research objective 2, it is essential to define the metrics that are widely recognized in the literature. By tracking the following metrics consistently throughout the experiments, a comprehensive and equitable comparison of the compression methods can be made, addressing all critical aspects:

1. **Test perplexity:** It serves as the primary metric for model performance and is analogous to an accuracy measure—the lower the perplexity, the better the performance. In the context of compression analysis, an ideal compression technique should yield minimal or no increase in test perplexity. Perplexity is calculated by exponentiating the cross-entropy loss on the test (or training/validation) set.

2. **Compression Ratio:** This metric quantifies the reduction in the volume of communicated gradient updates, with all compression methods aiming to maximize this ratio without compromising test perplexity. The compression ratio is a critical metric as it reflects the efficiency of each method in reducing communication overhead while maintaining model performance.

3. **Computation Time:** This metric reflects the impact of compression techniques on computational overhead. While such overhead is often considered negligible, it can become significant, particularly for more complex models. In distributed environments, where the primary goal is to reduce communication time, excessive computation time can negate the benefits of a high compression ratio. Thus, it is crucial to monitor this metric. In this study, computation time is measured by tracking the training time per epoch using the *time.process_time()* library method. As the workers are simulated in series, the recorded time is also divided by the number of workers.

4. **Communication Time:** It is calculated based on the size of the (compressed) gradients and an estimated Ethernet speed, assuming bi-directional communication through a parameter server.

5. **Convergence Rate:** This metric measures the number of epochs required for a model to converge. It is distinct from computation time because two methods may have the same per-epoch time yet differ in the number of epochs required to converge or trigger early stopping. This distinction is important for fair comparison, as certain compression methods—particularly those with computationally intensive sketching techniques—may have similar per-epoch costs but reach convergence faster.

## VI. RESULT ANALYSIS

### A. Baseline Analysis

Close attention is given to the results of the baseline experiments, which involve training the distributed system without any compression applied. These baseline results serve as a reference point for evaluating each performance metric. Additionally, the random-$k$ compression technique is employed to illustrate the effects of a naive compression approach relative to the baseline. As shown in Fig. 1, a substantial performance degradation is observed when using random-$k$, both with residual error correction (approximately 76%) and without (approximately 271%), positioning it as a worst-case technique for comparison purposes. In assessing the baseline performance as the number of workers increases, it is anticipated that the test perplexity would remain unchanged, given that distributed SGD should be equivalent to the single-machine case when no compression is applied. However, a slight discrepancy in performance is observed, which is attributed to floating point precision errors that accumulate during the training of models in PyTorch[9]. While exact test perplexity could theoretically be achieved by casting data to 64-bit double or long data types and truncating imprecision, this approach significantly impacted training time and reduced performance. When compression techniques are incorporated, these floating point errors are found to be negligible. This analysis highlights the sensitivity of distributed training systems to precision issues, especially when no compression is applied, and underscores the importance of carefully considering precision trade-offs when designing distributed models.

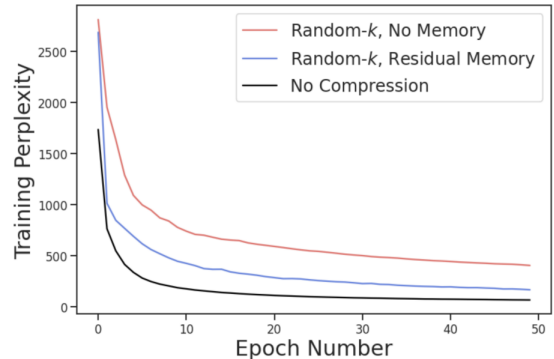

Fig. 1. Training ambiguity measured during the medium-sized model's one-worker training. The two compression methods that are used are random-$k$ with and without residual error correction and zero compression

### B. Compression vs Performance Analysis

For the medium-sized model, experiments were conducted using each compression technique across 1, 2, and 4 workers as depicted in Fig. 2. The results reveal a general trend of decreased model performance, consistent with hypothesis 1[10]. Interestingly, with 50× compression applied through top-$k$ and DGC, improved performance relative to the baseline was observed across all worker configurations. This represents the smallest degree of sparsification tested, in comparison to the 500× and 5000× compression ratios. These findings suggest that gradient sparsification, when applied conservatively, may have a regularizing effect on the model, whether trained in a distributed setting or otherwise. This result stands in contrast to the lower compression ratios achieved by QSGD, where model performance appears to degrade based on the compressed bit-rate. Hence, this potential regularization effect is not a function of compression ratio magnitude but rather of the specific technique employed. This observation underscores a key conclusion of this study, which is explored further in Section 6.5: fine-tuning models for each compression method individually can often result in performance improvements beyond those reported in the literature. For instance, a study on DGC with 462× compression reported a relative perplexity reduction of only 0.06. However, the baseline hyperparameters were retained in that study, which points to promising future research directions focused on separate tuning of models per compression method. Contrary to hypothesis 2[11], no consistent trend was observed in test performance as the number of workers increased. The hypothesis held true only in the case of zero compression as the model size changed. This suggests that scaling the number of workers should be used primarily as an indicator of training

---

[9] https://pytorch.org/

[10] An increased (worse) test perplexity relative to the baseline is expected to correspond with a rise in the compression ratio. Due to its momentum correcting residual memory, DGC is anticipated to outperform top-$k$ in the sparsification regime.

[11] Increasing the number of workers is expected to significantly decrease test performance in the end but have a good impact on calculation time regardless of compression method. This is because smaller mini-batches are used for the compression process.

run-time improvements, rather than as a method to enhance model performance.

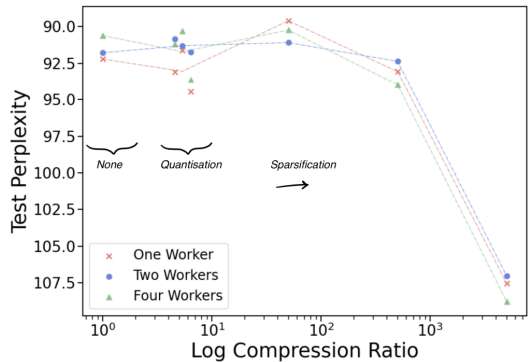

Fig. 2. Test the medium-sized model's perplexity vs. compression ratio for 1, 2, and 4 workers. For the sake of clarity, the results solely display top-$k$ for sparsification; a comparison between Top-k and DGC may be found in Fig. 4

Fig. 3 highlights the results of the sparsification regime. For lower levels of compression, DGC demonstrates better performance. However, at extremely high compression ratios (5000×), performance diverges significantly in favor of top-$k$. This is beyond the range typically reported for DGC (270×–600× compression). This finding suggests that while DGC excels in a lower sparsification regime, its performance suffers under extreme compression, despite the momentum factor designed to mitigate staleness. This implies that too many updates may have been masked at such high compression ratios.

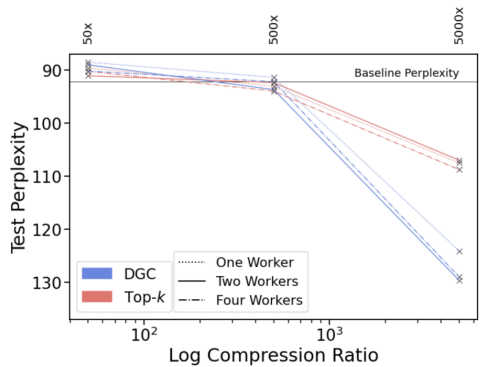

Fig. 3. Sparsification regime for top-$k$ and DGC in Fig. 5.2. Plots are made for tests with 1, 2, and 4 workers each

Fig. 4 offers another perspective on the trade-off between test perplexity and compression as model sizes vary in a single-worker setting. Once again, a regularizing effect is observed at 50× compression. However, as model size increases, the average improvement over the baseline diminishes. This suggests that the low-sparsification regularization effect is more prominent in smaller models. This finding implies that while performance improvements above the baseline are notable, they may be less relevant when considering that gradient compression is mainly applied to larger models. However, this auxiliary finding could inspire future research on the impact of gradient sparsification in improving the performance of smaller models, whether in language modeling or other domains. Furthermore, the anticipated performance degradation is observed with increasing compression ratios. Notably, this degradation is less pronounced in larger models, supporting hypothesis 3[12]. This suggests that the effectiveness of compression techniques, as reported in the literature, may also depend on the size of the model being tested. Specifically, it is possible to achieve negligible performance losses at higher compression levels for larger models, thereby allowing claims of superior technique effectiveness. In the quantization regime, the results are modest, with performance largely in line with baseline expectations. In the case of medium and larger models, 5-bit compression appears to perform best, though this conclusion is tentative, as differences in performance are likely attributable to the error tolerance afforded by the random search procedure.

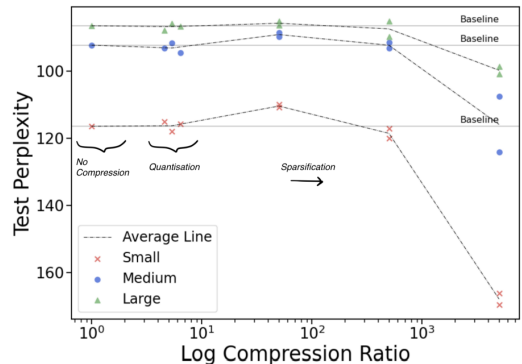

Fig. 4. Test the ratio of compression to perplexity in an environment with just one worker. Plotting was done by DGC and top-$k$ (not distinguishable). The model sizes are around 1, 4, and 9 million parameters, respectively, ranging from tiny to huge

## C. Computation vs Communication Analysis

Running times for the medium-sized model are compared across a range of worker configurations in Fig. 5. For each compression technique, both computation time and communication time are measured and stacked to provide a comprehensive overview of the total running time. Given the nature of simulating a distributed environment on a single machine, communication time is approximated as constant $n > 1$ workers. This simplification overlooks practical fluctuations and additional time incurred due to physical communication. Nonetheless, it is evident that for the baseline case, as the number of workers increases, communication time constitutes a larger proportion of the overall runtime of the model. This trend underscores the impetus behind the pursuit of communication-efficient training through gradient compression. It is observed that QSGD adds a modest amount of time to the training process, averaging 9% across all worker configurations. In contrast, sparsification techniques are predicted by hypothesis 4[13] to result in a significantly greater increase in computation time. However, this hypothesis is substantiated only in the case of top-$k$; DGC experiences an average increase of approximately 13% relative to the baseline computation, which is comparable to the increase observed with QSGD. Conversely, top-*k* exhibits an average increase of 26%, surpassing the total baseline running times in the four-worker scenario—the only instance where any of the compression techniques exceeded baseline run times. This increase can be attributed to the necessity of sorting entire tensor elements during compression, as opposed to DGC's approach, which samples only approximately 1% of the tensor prior to sorting. Specifically for top-*k*, runtime measurements are presented for the *torch.sort()* method, which has been reported in the literature to be faster than the standard *torch.topk()* method. This finding was confirmed during testing, revealing that the *torch.sort()* method is indeed approximately 20% faster. This highlights how implementation details can significantly impact the perceived effectiveness of a compression method. The low-opacity, hatched bars in Fig. 5 represent the simulated communication time for each compression technique, calculated based on bi-directional communication of model parameters over a 10 Gbps Ethernet connection. Given that this represents an estimation, additional factors such as headers and error corrections are excluded from the communication time calculations. This speed was selected to reflect the experimental conditions under which QSGD was studied, specifically utilizing Amazon EC2 pc2.16xlarge[14] instances. All compression methods effectively reduce communication times relative to the baseline, as illustrated in Fig. 5. At the scale of this study, the sparsification methods top-*k* and DGC reduce communication time to a negligible proportion of the overall runtime, even at compression ratios of 50×. This demonstrates the efficacy of each technique in mitigating this operational cost.

---

[12] Because there are more "redundant" model parameters when the size of the model increases, it is anticipated that the relative performance of high compression approaches will improve. In other words, high compression will become more effective as the model size increases.

[13] The sparsification techniques' computation time given that DGC and top-$k$ involve sorting tensor data, their anticipated values are significantly higher than those for quantization and no compression. This might offset the time saved from reduced communication.

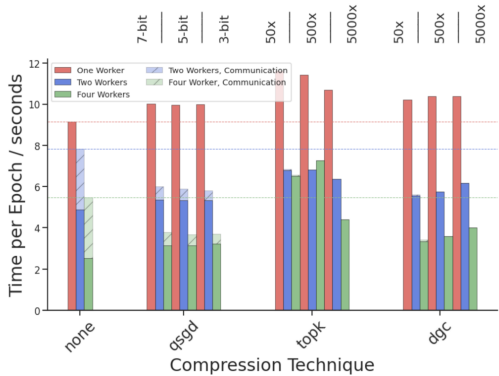

Fig. 5. Computation and communication times for every compression method and ratio examined throughout the medium-sized model's training

Additionally, the effect of increasing model size on computation time for distributed models employing different compression techniques is explored in Fig. 6. It is noted that, relative to the baseline, both QSGD and DGC techniques exhibit superior scalability compared to top-*k* in terms of additional computation time. Quantitatively, for the large model scenario, QSGD and DGC add approximately 3.3 seconds per epoch, whereas top-*k* necessitates an additional 5.8 seconds. Coupled with previous findings indicating the inferior scalability of top-*k* relative to other techniques, this positions top-*k* as a less favorable option for minimizing model training runtimes. Furthermore, minimal variation in computation time is observed as the compression ratio is altered. Conducting distributed training within a simulated, single-GPU environment presents certain limitations for runtime analysis. First, the choice of Ethernet speed for communication calculations is somewhat arbitrary and fails to capture the fluctuations inherent to physical systems. Second, the RAM of the T4 model utilized for computation is 16 GB, in contrast to the 732 GB of host memory employed in the larger-scale environments where QSGD was studied. This discrepancy significantly affects absolute computation times. Consequently, the computation-to-communication stacked bar charts presented herein are at an arbitrary ratio, intended to approximate physical settings as closely as possible. In reality, communication times are likely to be somewhat higher, and computation times could be lower with more efficient memory utilization. Thus, this subsection is intended to illustrate scaling trends and comparisons among compression techniques rather than to draw definitive conclusions.

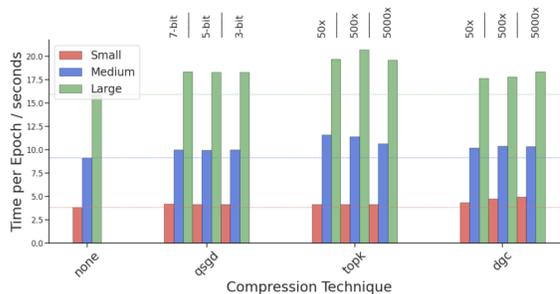

Fig. 6. Computation time for each compression method and ratio examined during the training of the small, medium, and big sized models. Plotted data for the scenario with a single worker

### D. Convergence Analysis

Fig. 7 presents a plot of the perplexity evaluated on the model throughout the training process for each compression method under investigation, utilizing a medium-sized model with a single worker. Notably, it is observed that, in comparison to the baseline case of zero compression, the sparsification techniques top-*k* and DGC exhibit an initial faster rate of convergence. In contrast, QSGD aligns more closely with the zero compression scenario. This finding suggests that, within a sparsification framework, a distributed model can achieve convergence—or a satisfactory approximation thereof—in a reduced number of epochs relative to a model trained without compression. This outcome has significant implications for computational efficiency, as analyzed in Section 6.3. The time savings associated with a model that converges in fewer epochs are likely to outweigh the savings incurred on a per-epoch basis. This enhanced convergence rate may be attributed to the generally lower learning rates identified as optimal for sparsification-based distributed models. This insight opens up intriguing possibilities for future research aimed at accelerating training processes through early stopping techniques applied to sparsified distributed models.

---

[14] https://aws.amazon.com/ec2/instance-types/p2/

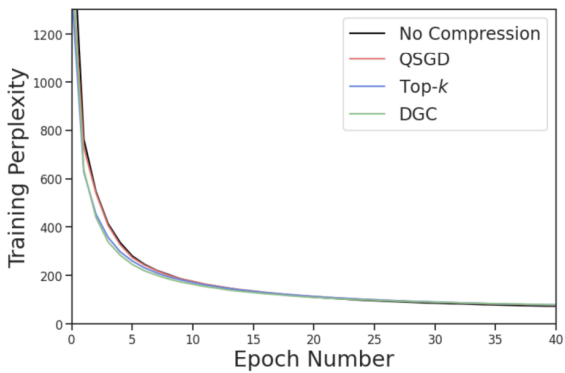

Fig. 7. Training ambiguity was monitored during the medium-sized model's one-worker training. Zero compression, QSGD, top-$k$, and DGC are the compression methods that are employed

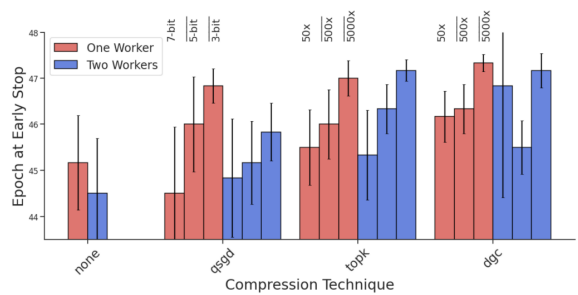

Fig. 8. Grouped bar graph showing, for both the 1 and 2 worker settings, the average epoch at which the medium size model converges when early halting is used. Trends were further categorized based on the compression method employed

To further examine this aspect of convergence, Fig. 8 presents a chart illustrating the epochs at which early stopping occurs, focusing on the fine-grained compression ratio options for each technique, using both one and two workers with the medium-sized model. It is important to note that this metric is somewhat compromised by a bug discovered late in the study, which caused the early stopping mechanism to increment for any non-improvement in validation perplexity during training, rather than adhering to a streak of non-improvements. While this issue is not anticipated to significantly affect performance results—given that the best-case epoch hovers around 45, suggesting a typical patience parameter of 5, which is not unusual—conclusions drawn regarding convergence variability for hypothesis 5[15] may be impacted by this oversight. Despite this limitation, a discernible trend emerges across all techniques, indicating that as compression increases, the average epoch at which early stopping occurs tends to be later. This result complements the previously identified accelerated convergence of sparsification-based distributed models, further underscoring the computational trade-offs associated with adopting higher compression ratios. Additionally, the graph incorporates standard mean error bars for the 'early-stopped epoch' metric. However, due to the constraints placed on the number of training epochs and the aforementioned bug in the early stopping implementation, no clear trend in variance is discernible. With increased resources, future work addressing this metric would be valuable for analyzing the performance-convergence variance trade-off claims associated with QSGD and for determining whether this theory holds true for sparsification techniques.

### E. Hyperparameter Choice Analysis

The research literature on gradient compression techniques, particularly those surveyed in this study, consistently maintains that the learning rate utilized during training remains aligned with the default baseline (zero compression) setting. Fig. 9 illustrates the distribution of learning rates observed for each compression technique across one, two, and four workers when the learning rate is specifically tuned. Notably, a marked difference in the magnitude of optimal learning rates emerges, particularly when comparing quantization to sparsification techniques. The top-$k$ and DGC methods typically adopt lower learning rates for optimal performance, in contrast to the higher values observed in the zero compression and QSGD methods. Furthermore, no discernible trend is apparent with regard to varying worker numbers. As referenced in Section 6.4, the adoption of lower learning rates may also correlate with the accelerated convergence characteristic of sparsification techniques—an aspect that has not been previously explored and which presents intriguing opportunities for optimization within compression techniques.

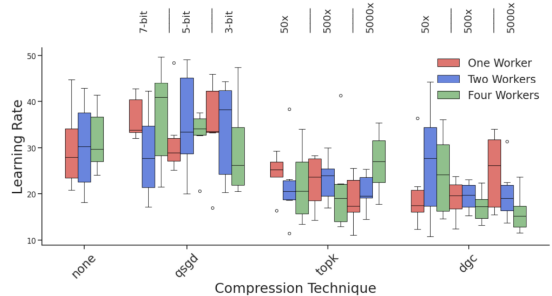

Fig. 9. A box plot that shows the range of learning rate values used across 1, 2, and 4 workers for each compression method. The medium model is the one that is visualized

In addition to learning rates, dropout values represent another hyperparameter warranting analysis. Fig. 10 visualizes the dropout values for one, two, and four workers utilizing the medium-sized model. While no clear trend emerges in QSGD as the compression ratio increases, a steady decline in optimal dropout values is observed within the sparsification regime, particularly evident in the two-worker case. This trend aligns with the rationale that fewer gradient elements are selected as the compression ratio increases, necessitating a lower dropout rate to prevent the exclusion of critical 'heavy-hitting' gradient elements. Moreover, it appears that the dropout rate

---

exhibits less sensitivity when no compression or QSGD is employed, suggesting that these approaches may offer greater stability during tuning. These findings underscore the necessity for meticulous tuning of distributed models to facilitate a fair analysis of compression techniques in relation to a baseline. In practical applications, the primary objective of compression techniques is to render the training of complex models feasible. Consequently, tuning would typically be performed on a model that incorporates gradient compression. By advocating for research that includes hyperparameter tuning in non-baseline models, the reported results can more accurately reflect practical scenarios, thereby affording a fairer opportunity for each technique to achieve optimal performance.

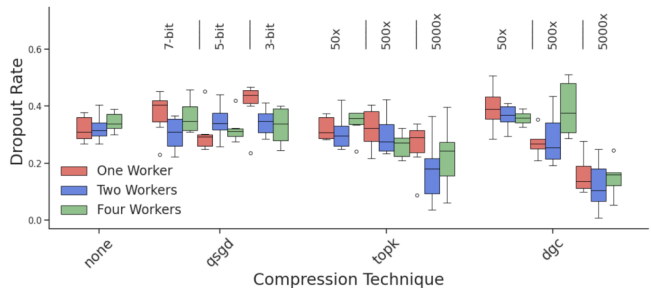

Fig. 10. A box plot that shows the range of dropout levels used for one, two, and four workers in each compression method. The medium model is the one that is visualized.

## VII. Conclusion and Future Works

This study investigates the impact of gradient compression techniques within the context of distributed DL, focusing on three prominent methods: top-$k$, DGC, and QSGD. The relevance of this investigation to language modeling is underscored by its low computation-to-communication ratio, which amplifies the issues associated with communication bottlenecks. The results demonstrate that, contrary to initial expectations, modest compression rates can yield performance improvements relative to a zero-compression baseline, particularly with the application of sparsification techniques. However, these enhancements tend to diminish as model size increases. While top-$k$ and DGC achieve higher compression rates, the additional computation time required for top-$k$ may offset the benefits gained from reduced communication time. The findings highlight the critical importance of hyperparameter tuning for each model type and the influence of various factors, including model size and available resources. Limitations arising from budgetary and resource constraints have implications for the scale and accuracy of the findings. Future research would benefit from more targeted investigations and the examination of a broader range of model types, especially as the scale of language models continues to expand.

## VIII. Declarations

*A.* **Funding:** No funds, grants, or other support was received.

*B.* **Conflict of Interest:** The authors declare that they have no known competing for financial interests or personal relationships that could have appeared to influence the work reported in this paper.

*C.* **Data Availability:** Data will be made on reasonable request.

*D.* **Code Availability:** Code will be made on reasonable request.

## References

[1] L. Wang *et al.*, "Document-Level Machine Translation with Large Language Models," Apr. 2023, Accessed: Aug. 23, 2023. [Online]. Available: https://arxiv.org/abs/2304.02210v1

[2] W. Xu, C. Napoles, E. Pavlick, Q. Chen, and C. Callison-Burch, "Optimizing Statistical Machine Translation for Text Simplification," *Trans. Assoc. Comput. Linguist.*, vol. 4, pp. 401–415, Dec. 2016, doi: 10.1162/tacl_a_00107.

[3] D. Bahdanau, K. H. Cho, and Y. Bengio, "Neural machine translation by jointly learning to align and translate," in *3rd International Conference on Learning Representations, ICLR 2015 - Conference Track Proceedings*, International Conference on Learning Representations, ICLR, Sep. 2015. Accessed: Oct. 21, 2023. [Online]. Available: https://arxiv.org/abs/1409.0473v7

[4] N. Kanda *et al.*, "Guided source separation meets a strong ASR backend: Hitachi/Paderborn university joint investigation for dinner party ASR," in *Proceedings of the Annual Conference of the International Speech Communication Association, INTERSPEECH*, International Speech Communication Association, May 2019, pp. 1248–1252. doi: 10.21437/Interspeech.2019-1167.

[5] J. Umamaheswari and A. Akila, "An Enhanced Human Speech Emotion Recognition Using Hybrid of PRNN and KNN," in *Proceedings of the International Conference on Machine Learning, Big Data, Cloud and Parallel Computing: Trends, Perspectives and Prospects, COMITCon 2019*, Institute of Electrical and Electronics Engineers Inc., Feb. 2019, pp. 177–183. doi: 10.1109/COMITCon.2019.8862221.

[6] B. Kang, Z. Liu, X. Wang, F. Yu, J. Feng, and T. Darrell, "Few-shot object detection via feature reweighting," in *Proceedings of the IEEE International Conference on Computer Vision*, 2019, pp. 8419–8428. doi: 10.1109/ICCV.2019.00851.

[7] J. N. Yasin, S. A. S. Mohamed, M. H. Haghbayan, J. Heikkonen, H. Tenhunen, and J. Plosila, "Low-cost ultrasonic based object detection and collision avoidance method for autonomous robots," *Int. J. Inf. Technol.*, vol. 13, no. 1, pp. 97–107, Feb. 2021, doi: 10.1007/s41870-020-00513-w.

[8] Y. Bang *et al.*, "A Multitask, Multilingual, Multimodal Evaluation of ChatGPT on Reasoning, Hallucination, and Interactivity," Feb. 2023, Accessed: Aug. 23, 2023. [Online]. Available: https://arxiv.org/abs/2302.04023v2

[9] D. Cai, W. Wang, and M. Li, "Incorporating Visual Information in Audio Based Self-Supervised Speaker Recognition," *IEEE/ACM Trans. Audio Speech Lang. Process.*, vol. 30, pp. 1422–1435, 2022, doi: 10.1109/TASLP.2022.3162078.

[10] Y. Lecun, Y. Bengio, and G. Hinton, "Deep learning," May 27, 2015, *Nature Publishing Group*. doi: 10.1038/nature14539.

[11] Ł. Kaiser and I. Sutskever, "Neural GPUs learn algorithms," in *4th International Conference on Learning Representations, ICLR 2016 - Conference Track Proceedings*, International Conference on Learning Representations, ICLR, Nov. 2016. Accessed: Jul. 20, 2023. [Online]. Available: https://arxiv.org/abs/1511.08228v3

[12] X. P. Qiu, T. X. Sun, Y. G. Xu, Y. F. Shao, N. Dai, and X. J. Huang, "Pre-trained models for natural language processing: A survey," Oct. 01, 2020, *Springer Verlag*. doi: 10.1007/s11431-020-1647-3.

[13] F. P. Such, V. Madhavan, E. Conti, J. Lehman, K. O. Stanley, and J. Clune, "Deep Neuroevolution: Genetic Algorithms Are a Competitive Alternative for Training Deep Neural Networks for Reinforcement Learning," Dec. 2017, doi: 10.48550/arxiv.1712.06567.

[14] F. Sattler, K. R. Muller, and W. Samek, "Clustered Federated Learning: Model-Agnostic Distributed Multitask Optimization under Privacy Constraints," *IEEE Trans. Neural Networks Learn. Syst.*, vol. 32, no. 8, pp. 3710–3722, Aug. 2021, doi:

10.1109/TNNLS.2020.3015958.

[15] L. Espeholt *et al.*, "IMPALA: Scalable Distributed Deep-RL with Importance Weighted Actor-Learner Architectures," in *35th International Conference on Machine Learning, ICML 2018*, PMLR, Jul. 2018, pp. 2263–2284. Accessed: Dec. 03, 2023. [Online]. Available: https://proceedings.mlr.press/v80/espeholt18a.html

[16] K. Fu, S. Kamara, and T. Kohno, "Key regression: Enabling efficient key distribution for secure distributed storage," *Comput. Sci. Dep. Fac. Publ. Ser.*, no. February, p. 149, 2006.

[17] Y. Zhan and J. Zhang, "An Incentive Mechanism Design for Efficient Edge Learning by Deep Reinforcement Learning Approach," in *Proceedings - IEEE INFOCOM*, Institute of Electrical and Electronics Engineers Inc., Jul. 2020, pp. 2489–2498. doi: 10.1109/INFOCOM41043.2020.9155268.

[18] Y. Lin, S. Han, H. Mao, Y. Wang, and W. J. Dally, "Deep Gradient Compression: Reducing the Communication Bandwidth for Distributed Training," *6th Int. Conf. Learn. Represent. ICLR 2018 - Conf. Track Proc.*, Dec. 2017, Accessed: Oct. 13, 2024. [Online]. Available: https://arxiv.org/abs/1712.01887v3

[19] J. Yang *et al.*, "Quantization networks," in *Proceedings of the IEEE Computer Society Conference on Computer Vision and Pattern Recognition*, 2019, pp. 7300–7308. doi: 10.1109/CVPR.2019.00748.

[20] F. Seide, H. Fu, J. Droppo, G. Li, and D. Yu, "1-bit stochastic gradient descent and its application to data-parallel distributed training of speech DNNs," in *Proceedings of the Annual Conference of the International Speech Communication Association, INTERSPEECH*, 2014, pp. 1058–1062. doi: 10.21437/interspeech.2014-274.

[21] D. Alistarh, D. Grubic, J. Z. Li, R. Tomioka, and M. Vojnovic, "QSGD: Communication-efficient SGD via gradient quantization and encoding," in *Advances in Neural Information Processing Systems*, 2017, pp. 1710–1721.

[22] P. Stone and M. Veloso, "Multiagent systems: a survey from a machine learning perspective," *Auton. Robots*, vol. 8, no. 3, pp. 345–383, Jun. 2000, doi: 10.1023/A:1008942012299.

[23] S. Dong, P. Wang, and K. Abbas, "A survey on deep learning and its applications," May 01, 2021, *Elsevier*. doi: 10.1016/j.cosrev.2021.100379.

[24] D. Feng and M. Q. Feng, "Computer vision for SHM of civil infrastructure: From dynamic response measurement to damage detection – A review," *Eng. Struct.*, vol. 156, pp. 105–117, Feb. 2018, doi: 10.1016/J.ENGSTRUCT.2017.11.018.

[25] C. Szegedy, V. Vanhoucke, S. Ioffe, J. Shlens, and Z. Wojna, "Rethinking the Inception Architecture for Computer Vision," in *Proceedings of the IEEE Computer Society Conference on Computer Vision and Pattern Recognition*, 2016, pp. 2818–2826. doi: 10.1109/CVPR.2016.308.

[26] H. Xu, *Lifelong Representation Learning for NLP Applications*. 2020. Accessed: Oct. 21, 2022. [Online]. Available: https://search.proquest.com/openview/8df895fd48ee44faf37ace0d22ba3fff/1?pq-origsite=gscholar&cbl=18750&diss=y

[27] V. K. Pallipuram, M. Bhuiyan, and M. C. Smith, "A comparative study of GPU programming models and architectures using neural networks," *J. Supercomput.*, vol. 61, no. 3, pp. 673–718, Sep. 2012, doi: 10.1007/s11227-011-0631-3.

[28] D. B. Skillicorn and D. Talia, "Models and Languages for Parallel Computation," *ACM Comput. Surv.*, vol. 30, no. 2, pp. 123–169, Jun. 1998, doi: 10.1145/280277.280278.

[29] S. Li *et al.*, "PyTorch Distributed: Experiences on Accelerating Data Parallel Training," *Proc. VLDB Endow.*, vol. 13, no. 12, pp. 3005–3018, Jun. 2020, doi: 10.14778/3415478.3415530.

[30] Z. Jia, M. Zaharia, and A. Aiken, "Beyond Data and Model Parallelism for Deep Neural Networks," *Proc. Mach. Learn. Syst.*, vol. 1, pp. 1–13, Apr. 2018, Accessed: Oct. 13, 2024. [Online]. Available: http://arxiv.org/abs/1807.05358

[31] S. Dramé-Maigné, M. Laurent, L. Castillo, and H. Ganem, "Centralized, Distributed, and Everything in between: Reviewing Access Control Solutions for the IoT," *ACM Comput. Surv.*, vol. 54, no. 7, Sep. 2022, doi: 10.1145/3465170.

[32] A. Sampath and C. Tripti, "Synchronization in distributed systems," in *Advances in Intelligent Systems and Computing*, Springer, Berlin, Heidelberg, 2012, pp. 417–424. doi: 10.1007/978-3-642-31513-8_43.

[33] S. H. Hashemi, S. A. Jyothi, and R. H. Campbell, "TicTac: Accelerating Distributed Deep Learning with Communication Scheduling," *Proc. Mach. Learn. Syst.*, vol. 1, pp. 418–430, Apr. 2018, Accessed: Oct. 13, 2024. [Online]. Available: https://github.com/xldrx/tictac

[34] J. Wangni, J. Liu, J. Wang, and T. Zhang, "Gradient sparsification for communication-efficient distributed optimization," in *Advances in Neural Information Processing Systems*, 2018, pp. 1299–1309.

[35] Y. Zhou, S. M. Moosavi-Dezfooli, N. M. Cheung, and P. Frossard, "Adaptive quantization for deep neural network," in *32nd AAAI Conference on Artificial Intelligence, AAAI 2018*, AAAI press, Apr. 2018, pp. 4596–4604. doi: 10.1609/aaai.v32i1.11623.

[36] L. Marage *et al.*, "Characterisation of a split gradient coil design induced systemic imaging artefact on 0.35 T MR-linac systems," *Phys. Med. Biol.*, vol. 68, no. 1, p. 01NT03, Dec. 2023, doi: 10.1088/1361-6560/aca876.

[37] J. Langford, L. Li, and T. Zhang, "Sparse online learning via truncated gradient," in *Advances in Neural Information Processing Systems 21 - Proceedings of the 2008 Conference*, 2009, pp. 905–912.

[38] S. U. Stich, J. B. Cordonnier, and M. Jaggi, "Sparsified SGD with memory," in *Advances in Neural Information Processing Systems*, 2018, pp. 4447–4458.

[39] C. Y. Chen, J. Choi, D. Brand, A. Agrawal, W. Zhang, and K. Gopalakrishnan, "AdaComp : Adaptive Residual Gradient Compression for Data-Parallel Distributed Training," *Proc. AAAI Conf. Artif. Intell.*, vol. 32, no. 1, pp. 2827–2835, Apr. 2018, doi: 10.1609/AAAI.V32I1.11728.

[40] Z. Tang, S. Shi, W. Wang, B. Li, and X. Chu, "Communication-Efficient Data Parallel Distributed Deep Learning: A Comprehensive Survey," vol. 1, 2023, doi: 10.1145/nnnnnnn.nnnnnnn.

[41] P. Kaur, G. S. Kashyap, A. Kumar, M. T. Nafis, S. Kumar, and V. Shokeen, "From Text to Transformation: A Comprehensive Review of Large Language Models' Versatility," Feb. 2024, Accessed: Mar. 21, 2024. [Online]. Available: https://arxiv.org/abs/2402.16142v1

[42] P. Veličković, "The resurgence of structure in deep neural networks," *Icml*, no. January, 2019, Accessed: May 09, 2024. [Online]. Available: https://www.repository.cam.ac.uk/items/f3ae9bb9-4263-4792-8734-f366bf3473a3

[43] S. Hochreiter, J. S.-N. Computation, and U. 1997, "Long short-term memory," *Neural Comput.*, vol. 9, no. 8, pp. 1735–1780, 1997, Accessed: Nov. 19, 2022. [Online]. Available: https://ieeexplore.ieee.org/abstract/document/6795963/

[44] W. Wen *et al.*, "TernGrad: Ternary gradients to reduce communication in distributed deep learning," in *Advances in Neural Information Processing Systems*, 2017, pp. 1510–1520. Accessed: Oct. 14, 2024. [Online]. Available: https://github.com/wenwei202/terngrad

[45] S. Zhou, Y. Wu, Z. Ni, X. Zhou, H. Wen, and Y. Zou, "DoReFa-Net: Training Low Bitwidth Convolutional Neural Networks with Low Bitwidth Gradients," Jun. 2016, Accessed: Oct. 14, 2024. [Online]. Available: https://arxiv.org/abs/1606.06160v3

[46] N. Strom, "Scalable distributed DNN training using commodity GPU cloud computing," in *Proceedings of the Annual Conference of the International Speech Communication Association, INTERSPEECH*, 2015, pp. 1488–1492. doi: 10.21437/interspeech.2015-354.

[47] N. Dryden, T. Moon, S. A. Jacobs, and B. Van Essen, "Communication Quantization for Data-Parallel Training of Deep Neural Networks," Institute of Electrical and Electronics Engineers (IEEE), Jan. 2017, pp. 1–8. doi: 10.1109/mlhpc.2016.004.

[48] A. F. Aji and K. Heafield, "Sparse communication for distributed gradient descent," in *EMNLP 2017 - Conference on Empirical Methods in Natural Language Processing, Proceedings*, Association for Computational Linguistics (ACL), Apr. 2017, pp. 440–445. doi: 10.18653/v1/d17-1045.

[49] H. Xu *et al.*, "Compressed Communication for Distributed Deep Learning: Survey and Quantitative Evaluation," *Uniw. śląski*, pp. 343–354, 2020, Accessed: Oct. 13, 2024. [Online]. Available: http://hdl.handle.net/10754/662495

[50] M. P. Marcus, "Building a large annotated corpus of English: the Penn Treebank (1993)," in *Corpus Linguistics: Readings in a Widening Discipline*, 2005, pp. 242–257.

[51] J. R. Hermans, G. Spanakis, and R. Möckel, "Accumulated gradient normalization," in *Journal of Machine Learning Research*, PMLR, Nov. 2017, pp. 439–454. Accessed: Oct. 13, 2024. [Online]. Available: https://proceedings.mlr.press/v77/hermans17a.html